\documentclass{article}

\PassOptionsToPackage{numbers, compress}{natbib}

     \usepackage[preprint]{neurips_2022}

\usepackage[utf8]{inputenc} 
\usepackage[T1]{fontenc}    
\usepackage{hyperref}       	
\usepackage{url}            	
\usepackage{booktabs}       	
\usepackage{amsfonts}       	
\usepackage{nicefrac}       	
\usepackage{microtype}      	
\usepackage{xcolor}            

\usepackage{epsfig}		
\usepackage{latexsym}		
\usepackage{amssymb}		

\newcommand{\boldu}{\mbox{\boldmath $u$}}

\newcommand{\boldw}{\mbox{\boldmath $w$}}

\newcommand{\boldW}{\mbox{\boldmath $W$}}
\newcommand{\boldX}{\mbox{\boldmath $X$}}
\newcommand{\boldY}{\mbox{\boldmath $Y$}}
\newcommand{\boldZ}{\mbox{\boldmath $Z$}}

\newcommand{\RR}{\mathbb{R}}

\newcommand{\boldmu}{\mbox{\boldmath $\mu$}}	
\newcommand{\boldTheta}{\mbox{\boldmath $\Theta$}}

\newcommand{\deff}{\stackrel{\rm def}{=}}
\newcommand{\noin}{\noindent}

\newtheorem{niteth}{Theorem} 		
\newtheorem{nitecoro}{Corollary} 		
\newcommand{\thend}{\hfill $\Box$}	
\newcommand{\qed}{\hbox{\rule{6pt}{6pt}}} 

\title{Spurious Local Minima of Deep ReLU Neural Networks 
in the Neural Tangent Kernel Regime}

\author{%
Tohru Nitta
\thanks{Part of this work has been done at the National Institute of 
Advanced Industrial Science and Technology.} \\
Graduate School of Artificial Intelligence and Science \\
Rikkyo University \\
Toshima-ku, Tokyo Japan \\
\texttt{tnitta@rikkyo.ac.jp} \\
}

\begin{document}

\maketitle

\begin{abstract}
In this paper, we theoretically prove that the deep ReLU neural networks 
do not lie in spurious local minima in the loss landscape 
under the Neural Tangent Kernel (NTK) regime, 
that is, in the gradient descent training dynamics of the deep ReLU neural networks 
whose parameters are initialized by a normal distribution 
in the limit as the widths of the hidden layers tend to infinity. 
\end{abstract}


\section{Introduction}
\label{intro}
Hinton et al. proposed Deep Belief Networks with a learning algorithm 
that trains one layer at a time \cite{hinton}. 
Since that report, deep neural networks have attracted attention extensively 
because of their human-like intelligence achieved through learning and generalization.
To date, deep neural networks have produced outstanding results 
in the fields of image processing, speech recognition and machine translation 
\cite{mohamed,seide,parcollet2019,taigman,sutskever}. 
Moreover, their scope of application has expanded, 
for example, to the field of mathematics \cite{davies}.

\noin
{\bf Local minima in neural networks.} 
On the one hand, local minima of neural networks have been investigated for a long time. 
Local minima cause plateaus which have a strong 
negative influence on learning in neural networks \cite{amari,cous}. 
%
Fukumizu et al. have mathematically proved that critical points introduced 
by a hierarchical structure in a three-layered neural network can be local minima 
or saddle points according to conditions \cite{fuku2000}. 
%
Dauphin et al. experimentally investigated the distribution of the critical points 
of a single-layer MLP and demonstrated that the possibility of existence of local 
minima with large error (i.e., bad or spurious local minima) is very small \cite{dauphin}. 
%
Yun et al. investigated the loss surface of three-layered neural networks with 
the standard activation functions such as ReLU, Leaky ReLU, sigmoid, $\tanh, 
\arctan$, ELU and SELU, and constructively proved that there is a spurious 
local minimum \cite{yun}. 
%
As for deep neural networks, 
Choromanska et al. provided a theoretical justification for the work in  
\cite{dauphin} on a deep neural network with ReLU units 
using the spherical spin-glass model under seven assumptions 
\cite{choroman-henaff}. 
%
Choromanska et al. also suggested that discarding the seven unrealistic assumptions 
remains an important open problem 
\cite{choroman-lecun}. 
%
Kawaguchi discarded most of these assumptions and 
proved that the following four statements for a deep ReLU neural network 
with only two out of the seven assumptions \cite{kawaguchi}: 
1) the loss function is non-convex and non-concave, 
2) every local minimum is a global minimum, 
3) every critical point that is not a global minimum is a saddle point, and 
4) bad saddle points exist. 
%
Nitta showed that 
there exist a large number of critical points introduced by a hierarchical structure 
in deep neural networks as straight lines, and derived a sufficient condition for deep neural 
networks having no critical points introduced by a hierarchical structure \cite{nit2017}. 
%
Laurent et al. studied the loss surface of deep ReLU or Leaky ReLU neural networks 
applied to classification problems, and proved that there are only two types 
of local minima: flat minima and sharp minima. 
They clarified that the sharp minima are spurious local minima, 
and that the flat minima are global minima in the case of the Leaky ReLU network 
\cite{laurent}. 
%
Liu et al. constructively proved that there exist non-differentiable saddle points 
in the loss surface of deep ReLU networks with squared loss or cross-entropy loss, 
and that deep ReLU networks with cross-entropy loss have non-differentiable 
spurious local minima under a mild restriction on datasets \cite{Liu}. 
%
Incidentally, Nitta investigated the characteristics of the complex-valued neuron 
model with parameters represented by polar coordinates and showed that 
singular points degrade the learning speed in the case of using the steepest gradient 
descent method with square error \cite{nit2015}. 
As for three-layered complex-valued neural networks, 
Nitta mathematically proved that most of local minima caused by the hierarchical 
structure can be resolved by extending the real-valued neural network to 
complex numbers \cite{nit2013}.

\noin
{\bf Neural Tangent Kernel.} 
On the other hand, it is proved that for a least-squares regression cost, 
the vector of the parameters of deep neural networks has the same distribution 
as the initial value of the vector of the parameters 
for all times in the Neural Tangent Kernel (NTK) regime 
where the vector of the parameters is initialized with a normal distribution, 
and the width of hidden layers is infinity \cite{jacot}. 
Since then, theoretical research in the NTK regime has become active. 
%
Allen-Zhu et al. proved that the parameters of deep ReLU neural networks move little 
from their initial values during training such as stochastic gradient descent 
and used it to show the training algorithm can find global minima on the 
error function in polynomial time in the NTK regime \cite{zhu}. 
%
Du et al. proved the same thing as in \cite{zhu} for the case of Lipschitz 
and smooth activation functions such as the soft-plus and sigmoid \cite{du}. 
%
Lee et al. theoretically showed that the learning dynamics with 
a certain learning rate 
in parameter space of deep nonlinear neural networks are exactly described 
by a linearized model where the parameters of the deep neural networks 
move little from their initial values in the NTK regime \cite{lee}. 
That is, the trained network is equivalent to a linearized model around random initialization. 
%
Karakida et al. analyzed the generalization performance of continual learning 
in the NTK regime \cite{karakida}. 

\noin
{\bf Results of this paper.} 
In this paper, by discarding the remaining two assumptions in \cite{kawaguchi}, 
we prove that 
the deep ReLU neural networks 
do not lie in spurious local minima in the loss landscape 
under the NTK regime, 
that is, in the gradient descent training dynamics of the deep ReLU neural networks 
whose parameters are initialized by a normal distribution 
in the limit as the widths of the hidden layers tend to infinity.

\section{Analysis on spurious local minima}
\label{analysis_minima}

In this section, we theoretically prove that the deep ReLU neural networks 
do not lie in spurious local minima in the loss landscape 
under the NTK regime.

\subsection{Kawaguchi model}
\label{kawa-model}

This subsection presents a description of the deep ReLU neural network 
model analyzed in \cite{kawaguchi} (we call it {\it Kawaguchi model} here).

First, we consider the following neuron. 
The net input $U_n$ to a neuron $n$ is defined as:  
$U_n = \sum_mW_{nm}I_m$, 
where $W_{nm}$ represents the weight connecting the neurons $n$ and $m$, 
$I_m$ represents the input signal from the neuron $m$. 
It is noteworthy that biases are omitted for the sake of simplicity. 
The output signal is defined as $\varphi(U_n)$ 
where $\varphi(u) \deff \max(0, u)$ for any $u \in \RR$ and is called {\it Rectified Linear Unit} 
({\it ReLU}, $\RR$ \ denotes the set of real numbers). 

\begin{figure*}
\centerline{\epsfig{file=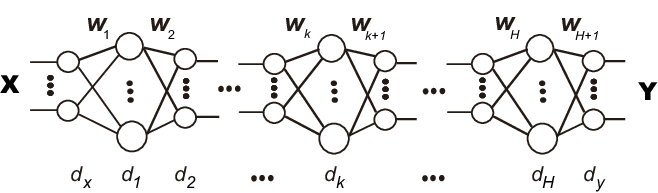,width=14cm,height=4cm}}
\caption{A deep neural network model \cite{kawaguchi}: 
$(\boldX, \boldY)$ are the training data, 
$\boldW_k$ signifies the weight matrix between the $(k-1)$-th layer and the 
$k$-th layer, and $d_k$ denotes the number of neurons of the $k$-th layer.}
\label{fig1}
\end{figure*}

The deep nonlinear neural network described in \cite{kawaguchi} 
consists of such neurons described above (Fig.~\ref{fig1}). 
The network has $H+2$ layers ($H$ is the number of hidden layers). 
The activation function $\psi$ of the neuron in the output layer is linear, 
i.e., $\psi(u)=u$ for any $u \in \RR$.  
For any $0 \leq k \leq H+1$, let 
$d_k$ denote the number of neurons of the $k$-th layer, that is, the width 
of the $k$-th layer where the 0-th layer is the input layer and the 
$(H+1)$-th layer is the output layer. 
Let $d_x = d_0$ and $d_y = d_{H+1}$ for simplicity.

Let $(\boldX, \boldY)$ be the training data where $\boldX \in \RR^{d_x \times m}$ 
and $\boldY \in \RR^{d_y \times m}$ and where $m$ denotes the number of training patterns. 
We can rewrite the $m$ training data as $\{ (\boldX_i, \boldY_i) \}_{i=1}^m$ where 
$\boldX_i \in \RR^{d_x}$ is the $i$-th input training pattern and 
$\boldY_i \in \RR^{d_y}$ is the $i$-th output training pattern. 
Let $\boldW_k$ denote the weight matrix between 
the $(k-1)$-th layer and the $k$-th layer for any $1 \leq k \leq H+1$. 
Let $\boldTheta$ denote the one-dimensional vector which consists 
of all the weight parameters of the deep nonlinear neural network.

\begin{figure*}
\centerline{\epsfig{file=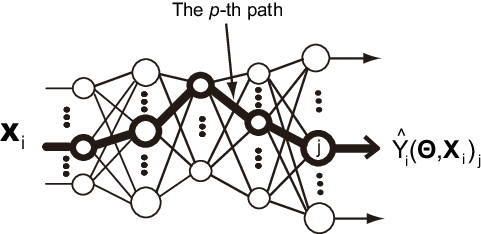,width=12cm,height=4cm}}
\caption{Image of a path in the deep neural network model \cite{kawaguchi}: 
$\boldX_i$ denotes the $i$-th input training pattern, and 
$\hat{\boldY}_i(\boldTheta, \boldX_i)_j$ stands for the actual output of output neuron $j$. 
}
\label{fig2}
\end{figure*}

Kawaguchi specifically examined a path from an input neuron to an output neuron of 
the deep ReLU neural network (Fig.~\ref{fig2}), and expressed 
the actual output of output neuron $j$ of the output layer of the deep ReLU 
neural network %
for the $i$-th input training pattern $\boldX_i \in \RR^{d_x}$ as 
\begin{eqnarray}
\label{eqn2-1}
\hat{\boldY}_i(\boldTheta, \boldX_i)_j = q\sum_{p=1}^\Psi [\boldX_i]_{(j,p)} [\boldZ_i]_{(j,p)} 
\prod_{k=1}^{H+1} w_{(j,p)}^{(k)} \in \RR 
\end{eqnarray}
where $\Psi$ represents the total number of paths from the input layer 
to output neuron $j$, 
$[\boldX_i]_{(j,p)} \in \RR$ denotes the component of the $i$-th input training pattern 
$\boldX_i \in \RR^{d_x}$  
that is used in the $p$-th path to the $j$-th output neuron, 
and $q$ is a constant for normalization. 
Also, $[\boldZ_i]_{(j,p)} \in \{0, 1\}$ represents whether the $p$-th path to the 
output neuron $j$ is active or not 
for each training pattern $i$ as a result of ReLU activation. 
$[\boldZ_i]_{(j,p)} = 1$ means that the path is active, and 
$[\boldZ_i]_{(j,p)} = 0$ means that the path is inactive. 
$w_{(j,p)}^{(k)} \in \RR$ is the component of the weight matrix 
$\boldW_k \in \RR^{d_k \times d_{k-1} }$ 
that is used in the $p$-th path to the output neuron $j$. 

The objective of the training is to find the parameters which minimize 
the error function defined as 
\begin{equation}
\label{eqn2-2}
L(\boldTheta) = \frac{1}{2} \sum_{i=1}^m 
E_{\boldZ} \Vert \hat{\boldY}_i(\boldTheta, \boldX_i) - \boldY_i \Vert^2 
\end{equation}
where $\Vert \cdot \Vert$ is the Euclidean norm, that is, 
$\Vert \boldu \Vert = \sqrt{u_1^2 + \cdots + u_N^2}$ 
for a vector $\boldu = (u_1 \cdots u_N)^T \in \RR^N$, 
and $\hat{\boldY}_i(\boldTheta, \boldX_i) \in \RR^{d_y}$ is the actual output of the output 
layer of the deep nonlinear neural network for the $i$-th training pattern $\boldX_i$. 
The expectation in Eq. (\ref{eqn2-2}) is made with respect to random vector 
$\boldZ = \{ [\boldZ_i]_{(j,p)} \}$.

The Kawaguchi model has been analyzed based on the following two assumptions. 

\noin 
{\bf A1p-m} 
{\it 
$P([\boldZ_i]_{(j,p)} = 1) = \rho$ for all $i$ and $(j,p)$ where $\rho \in \RR$ is a constant. 
That is, $[\boldZ_i]_{(j,p)}$ is a Bernoulli random variable.} 

\noin 
{\bf A5u-m} 
{\it 
$\boldZ$ is independent of the input $\boldX$ and the parameter $\boldTheta$.}

A1p-m and A5u-m are weaker ones of the two assumptions A1p and A5u 
in \cite{choroman-henaff}, 
respectively. 
The next corollary is one of the main results in \cite{kawaguchi}. 


\begin{nitecoro}(Kawaguchi, 2016: deep ReLU networks)
\label{col-kawa}
Assume A1p-m and A5u-m. Let $q=\rho^{-1}$. 
Further, assume that $\boldX\boldX^T$ and $\boldX\boldY^T$ are full rank, 
and $d_y \leq d_x$. Then, for any depth $H \geq 1$ 
and for any layer widths and any input-output dimensions $d_y, d_H, d_{H-1}, 
\cdots, d_1, d_x \geq 1$, the error function $L(\boldTheta)$ has the 
following properties: 
\begin{enumerate}
\item It is non-convex and non-concave. 
\item Every local minimum is a global minimum. 
\item Every critical point that is not a global minimum is a saddle point. 
\item If rank $(W_H \cdots W_2)=\min(d_H, \cdots, d_1)$, 
then the Hessian at any saddle point 
has at least one (strictly) negative eigenvalue. 
\end{enumerate}
\thend
\end{nitecoro}

Note that the assumptions on $\boldX\boldX^T$ and $\boldX\boldY^T$ 
in Corollary \ref{col-kawa} are realistic and easy to satisfy.

Strictly speaking, the following assumption A5u-m-1 suffices for the proof instead of 
the assumption A5u-m described above   

\noin 
{\bf A5u-m-1} 
{\it 
For any $i$ and any $(j,p)$, $[\boldZ_i]_{(j,p)} \in \{0, 1\}$ is independent of 
the $i$-th input training pattern 
$\boldX_i \in \RR^{d_x}$ and the sequence of the weights on the $p$-th path 
$\{ w_{(j,p)}^{(k)} \in \RR \}_{k=1}^{H+1}$ where $w_{(j,p)}^{(k)}$ is the weight 
between the layer $k-1$ 
and the layer $k$ on the $p$-th path ($k=1, \cdots, H+1$). }

Actually, according to assumption A5u-m-1, 
\begin{eqnarray}
\label{eqn2-3}
& & E_{\boldZ} \left[\hat{\boldY}_i(\boldTheta, \boldX_i)_j \right] \nonumber\\
&=& E_{\boldZ} \left[q\sum_{p=1}^\Psi [\boldX_i]_{(j,p)} [\boldZ_i]_{(j,p)} 
\prod_{k=1}^{H+1} w_{(j,p)}^{(k)} \right] 
\hspace*{1cm} \mbox{(from Eq.} (\ref{eqn2-1})) \nonumber\\
&=& q\sum_{p=1}^\Psi [\boldX_i]_{(j,p)} E_{\boldZ} \left[ [\boldZ_i]_{(j,p)} \right] 
\prod_{k=1}^{H+1} w_{(j,p)}^{(k)}.  
\hspace*{1cm} \mbox{(from the assumption A5u-m-1) }
\end{eqnarray}

Kawaguchi analyzed the deep ReLU neural network using Eq. (\ref{eqn2-3}) 
(see Section 3.2 in \cite{kawaguchi}).

\subsection{Analysis}
\label{analysis}
This subsection presents an analysis of the Kawaguchi model in the NTK regime. 


\noin
{\bf NTK regime.} 
First, we here summarize the NTK regime. 
Consider a fully connected deep neural network defined as 
\begin{eqnarray}
\label{eqn-intro-1}
h_l &=& \phi(u_l),\\
u_l &=& \frac{\sigma_w W_l h_{l-1} }{\sqrt{M_{l-1} }} + \sigma_b b_l \hspace{1cm} 
(l = 1, \cdots, L)
\end{eqnarray}
where $\phi$ is an activation function, $W_l \in \mathbb{R}^{{M_l}\times M_{l-1}}$ 
the weight matrix between the layer $l-1$ and layer $l$, $b_l \in \mathbb{R}^{M_l}$ 
the bias of the layer $l$, 
$\sigma_w^2$ the variance of weights, $\sigma_b^2$ the variance of biases,  
and $M_l$ the number of neurons in the hidden layer $l$. 
Each element of the weight and the bias is initialized according to the 
normal distribution $N(0, 1)$. 
That is, we can assume that each weight is initialized according to the 
normal distribution $N(0, \sigma_w^2/M_{l-1})$. 
Then one discusses the behavior or properties of the deep neural network when 
$M_l \rightarrow \infty$, that is, the width of hidden layers goes to infinity. 
This is known as the NTK regime. 


\newpage 

\noin
{\bf Kawaguchi model in the NTK regime.} 
We assume in this analysis that each weight between the layer $l$ and the layer $l+1$ 
is set according to the normal distribution $N(0, \sigma_l^2)$ 
where $\sigma_l = 1/(3d_l)$ and $d_l$ is the number of neurons in the layer $l$. 
We also assume that the width of the deep ReLU neural network is sufficiently large, 
that is, $d_0 (=d_x), d_1, \cdots, d_{H-1}$ are sufficiently large (Fig.~\ref{fig1}), and that 
each element of the $i$-th input training pattern $\boldX_i$ takes a value between 
$-\alpha$ and $\alpha$, that is, $\boldX_i \in I^{d_x}$ for any $i$ 
where $I=[-\alpha, \alpha]$ and $\alpha$ is a positive real number. 
By adding these assumptions described above, we can regard that the 
Kawaguchi model is in the NTK regime, 
and call it the {\it Kawaguchi model in the NTK regime}.

Incidentally, in He initialization which is commonly used in practice, 
the initial value of each weight between the layer $l$ 
and the layer $l+1$ is set according to either of the normal distributions 
$N ( 0, 2 / d_l )$ or $N ( 0, 2 / d_{l+1} )$ independently \cite{he}. 
Thus, the case of the He initialization is included in the Kawaguchi 
model in the NTK regime.

\begin{figure*}
\centerline{\epsfig{file=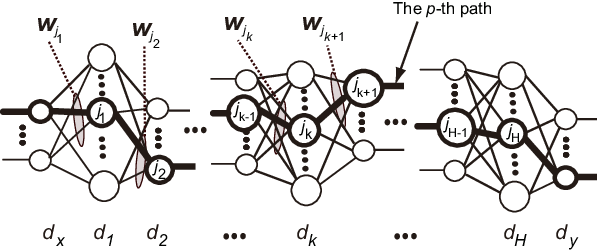,width=14cm,height=5cm}}
\caption{A deep ReLU neural network with $H$ hidden layers where 
$\boldw_{j_k}$ is the weight vector of hidden neuron $j_k$, 
hidden neurons $j_1, \cdots, j_H$ are on the $p$-th path, 
$d_k$ is the number of neurons of the $k$-th layer, 
$d_x = d_0$ and $d_y = d_{H+1}$. 
}
\label{fig4}
\end{figure*}

\noin
{\bf Flow of analysis.} 
We prove in the following Theorems \ref{thm1}-\ref{thm3} 
that the two assumptions A1p-m and A5u-m (A5u-m-1) 
are satisfied in the initial state immediately after the parameter initialization 
of the Kawaguchi model in the NTK regime. 
Then, we will realize in Theorem \ref{thm4} 
that the error function $L(\boldTheta)$ does not lie in spurious local minima 
in the initial state immediately after the parameter initialization 
under the NTK regime according to Corollary \ref{col-kawa}. 
After that, we will prove in Theorem \ref{thm5} that the 
error function $L(\boldTheta)$ does not lie in spurious local minima 
{\it during training} in the Kawaguchi model in the NTK regime.

\noin
{\bf Analysis in the initial state.}  
\begin{niteth}
\label{thm1}
For any training pattern $i$, any output neuron $j$, and any path $p$ 
from an input neuron to the output neuron $j$ 
in the initial state immediately after the parameter initialization 
of the Kawaguchi model in the NTK regime, 
\begin{equation}
\label{eqn4-1}
P\left([Z_i]_{(j,p)} = 1 \right) = \frac{1}{2^H}
\end{equation}
where $H$ is the number of hidden layers ($H \geq 1$). 
\thend
\end{niteth}

\begin{figure}[b]
\centerline{\epsfig{file=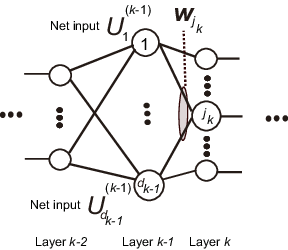,width=8cm,height=5.5cm}}
\caption{Relationship between the weight vector $\boldw_{j_k}$ and 
the net inputs $U_1^{(k-1)}, \cdots, U_{d_{k-1}}^{(k-1)}$. }
\label{fig5}
\end{figure}

\noin 
{\it Proof}. \ 
Denote by $j_1, \cdots, j_H$ the hidden neurons on path $p$ 
where $j_k$ is the hidden neuron in the $k$-th hidden layer $(k=1, \cdots, H$) 
(Fig.~\ref{fig4}). 
Then, 
\begin{eqnarray}
\label{eqn4-2}
P\Big( [Z_i]_{(j,p)} = 1 \Big) 
&=& P\Big( \mbox{Net input to the hidden neuron}\ j_k > 0 \ \ 
(k=1, \cdots, H) \Big) \nonumber\\
&=& P \left( \boldX_i^T \boldw_{j_1} > 0, \left[ \varphi( U_1^{(1)} ) \cdots \varphi(U_{d_1}^{(1)} ) \right] 
\boldw_{j_2} > 0, \right. \nonumber\\ 
& & \left. 
\cdots, \left[ \varphi( U_1^{(H-1)} ) \cdots \varphi(U_{d_{H-1}}^{(H-1)}) \right] 
\boldw_{j_H} > 0 \right)
\end{eqnarray}
where $\boldX_i \in I^{d_x}$ is the $i$-th input training pattern, 
$\boldw_{j_k} \in \RR^{d_{k-1}}$ is the weight vector of the hidden neuron $j_k$ 
in the hidden layer $k$, and 
$U_l^{(k)}$ is the net input to the hidden neuron $l$ in the hidden layer $k$ 
(Fig.~\ref{fig5}).

We prove by mathematical induction that Eq. (\ref{eqn4-1}) holds true. 

\begin{figure}
\centerline{\epsfig{file=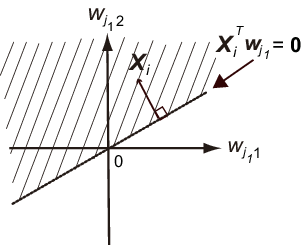,width=7cm,height=5.5cm}}
\caption{A half-open hyperspace 
\{$\boldw_{j_1} = (w_{{j_1} 1} \ w_{{j_1} 2})^T \in \RR^{d_x} \ \vert \ \boldX_i^T \boldw_{j_1} > 0$\} 
( a half-open plane because $d_x = 2$). }
\label{fig6}
\end{figure}

\noin
[For $H=1$] 
This case corresponds to a three-layered neural network. 
It follows that 
\begin{eqnarray}
\label{eqn4-3}
P\Big( [Z_i]_{(j,p)} = 1 \Big) &=& P ( \boldX_i^T \boldw_{j_1} > 0 ) \ \ \ 
(\mbox{from Eq.} (\ref{eqn4-2})) \nonumber\\
&=& \frac{1}{2} 
\end{eqnarray}
where $j_1$ is a hidden neuron on path $p$. 
We can see below that the last equality of Eq. (\ref{eqn4-3}) holds true. 
For a given input training pattern $\boldX_i \in I^{d_x}$, 
$\{ \boldw_{j_1} \in \RR^{d_x} \vert \ \boldX_i^T \boldw_{j_1} > 0 \}$ 
is a half-open hyperspace with a normal vector $\boldX_i$ through the origin 
in a $d_x$-dimensional Euclidean space (Fig.~\ref{fig6}). 
According to the assumption, 
the random variables $w_{j_11}, \cdots, w_{j_1d_x}$ which are the components of 
the weight vector $\boldw_{j1} = (w_{j_11} \cdots w_{j_1d_x})^T$ 
of the hidden neuron $j_1$ obey the normal distribution $( N, \sigma_x^2 )$ independently. 
Consequently, $P\left(\boldX_i^T \boldw_{j_1} > 0 \right) = P\left(\boldX_i^T \boldw_{j_1} \leq 0 \right)$, 
which means $P\left(\boldX_i^T \boldw_{j_1} > 0 \right) = 1/2$. 

\noin
[For $H$] 
This case corresponding to a deep ReLU neural network with $H$ hidden layers, 
we show that if the case of $H-1$ holds, then case $H$ also holds. 
Assuming that the case of $H-1$ holds, then
\begin{eqnarray}
\label{eqn4-4}
& &P\Big( [Z_i]_{(j,p)} = 1 \Big) \nonumber \\
&=& P \left( \boldX_i^T \boldw_{j_1} > 0, \left[ \varphi( U_1^{(1)} ) \cdots \varphi(U_{d_1}^{(1)} ) \right] 
\boldw_{j_2} > 0, \right. \nonumber\\ 
& & \left. \cdots, \left[ \varphi( U_1^{(H-1)} ) \cdots \varphi(U_{d_{H-1}}^{(H-1)}) \right] 
\boldw_{j_H} > 0 \right) \ \ 
(\mbox{Eq.} (\ref{eqn4-2}) \ \mbox{itself}) \nonumber\\
&=& P \left( \left[ \varphi( U_1^{(1)} ) \cdots \varphi(U_{d_1}^{(1)} ) \right] 
\boldw_{j_2} > 0, \cdots, 
%
\left[ \varphi( U_1^{(H-1)} ) \cdots \varphi(U_{d_{H-1}}^{(H-1)}) \right] 
\boldw_{j_H} > 0 \right. \nonumber\\
& & \hspace*{2cm} \left. \Big\vert \ \boldX_i^T \boldw_{j_1} > 0 \right) 
\cdot P\left( \boldX_i^T \boldw_{j_1} > 0 \right). 
\end{eqnarray}
Here, the first factor of the right-hand-side of Eq. (\ref{eqn4-4}) represents the 
probability that 
the path passing through the $H-1$ hidden neurons $j_2, \cdots, j_H$ 
for the input training pattern $[ \varphi( U_1^{(1)} ) \cdots \varphi(U_{d_1}^{(1)} ) ]^T \in \RR^{d_1}$ 
such that $U_{j_1}^{(1)} = \boldX_i^T \boldw_{j_1} > 0$ is active. 
%
Also, for any $1 \leq s \leq d_1$, by denoting $\boldX_i = (x_1 \cdots x_{d_0})^T$ 
and $\boldw_s = (w_1 \cdots w_{d_0})$ for the sake of simplicity, 
$P(\vert \varphi(U_s^{(1)}) \vert \leq \vert U_s^{(1)} \vert 
=\vert \boldX_i^T \boldw_s \vert 
\leq \alpha \sum_{t=1}^{d_0} \vert w_t \vert 
\leq 3 \alpha d_0 \sigma_0 = \alpha) = 0.9973 \fallingdotseq 1.0$ 
because each weight $w_t$ obeys the normal distribution $N(0, \sigma_0^2)$ 
where $\sigma_0 = 1/(3d_0)$ 
which means $P(\vert w_t \vert \leq 3 \sigma_0 = 1/d_0) = 0.9973$ 
(the so-called three-sigma rule of thumb). 
%
Hence, according to the assumption of mathematical induction, 
the first factor of the right-hand-side of Eq. (\ref{eqn4-4}) is equal to $(1/2)^{H-1}$. 
In addition, the second factor of the right-hand-side of Eq. (\ref{eqn4-4}) is equal to 1/2 
from Eq.(\ref{eqn4-3}). 
Therefore, 
\begin{eqnarray}
\label{eqn4-5}
P\Big( [Z_i]_{(j,p)} = 1 \Big) &=& \left(\frac{1}{2}\right)^{H-1} \cdot \frac{1}{2} \nonumber\\
&=& \frac{1}{2^H}, 
\end{eqnarray}
which means that the case of $H$ indeed holds. 
Therefore, by mathematical induction, Eq. (\ref{eqn4-1}) holds for any $H \geq 1$. 
\qed 

Theorem \ref{thm1} states that assumption A1p-m holds: $\rho=1/2^H$ in this case. 
Because 
\begin{eqnarray}
\lim_{H \rightarrow +\infty} P \left( [Z_i]_{(j,p)} = 1 \right) 
&=& \lim_{H \rightarrow +\infty} \frac{1}{2^H} \nonumber\\
&=& 0, 
\end{eqnarray}
the probability that path $p$ is active decreases exponentially. 
It converges to zero as the number of hidden layers $H$ increases. 
The probability that path $p$ is active decreases by half when a hidden layer is added.

\begin{niteth}
\label{thm2}
For any training pattern $i$, any output neuron $j$, and any path $p$ 
from an input neuron to the output neuron $j$ 
in the initial state immediately after the parameter initialization 
of the Kawaguchi model in the NTK regime, 
random variable $[Z_i]_{(j,p)}$ is independent of 
the sequence of weights on path $p$. 
\thend
\end{niteth}

At first glance, Theorem \ref{thm2} seems counterintuitive 
because the random variable $[Z_i]_{(j,p)}$ is a function of 
the sequence of weights on path $p$ 
in the model with the finite widths of the hidden layers. 
However, Theorem \ref{thm2} holds true because we deal with the deep ReLU 
neural network in the almost infinite-width limit. 
Actually, in the almost infinite-width limit, 
the absolute value of the product of an input value and the weight value  
on the path is negligibly small 
(see Eq. (\ref{eqn4-7-1}) in the supplementary material for details). 

The reason why the next theorem \ref{thm3} holds true is also as described above.

\begin{niteth}
\label{thm3}
For any training pattern $i$, any output neuron $j$, and any path $p$ 
from an input neuron to the output neuron $j$ 
in the initial state immediately after the parameter initialization 
of the Kawaguchi model in the NTK regime, 
random variable $[Z_i]_{(j,p)}$ is independent of 
the input training singnal $\boldX_i$. 
\thend
\end{niteth}

The proofs of Theorem \ref{thm2} and Theorem \ref{thm3} are 
in the supplementary material.

Theorem \ref{thm2} and Theorem \ref{thm3} state that assumption A5u-m-1 
holds in the initial state immediately after the parameter initialization 
of the Kawaguchi model in the NTK regime.

Thus, it follows from Theorems \ref{thm1}-\ref{thm3} that assumptions 
A1p-m and A5u-m (A5u-m-1) hold 
in the initial state immediately after the parameter initialization 
of the Kawaguchi model in the NTK regime, 
both of which were introduced in \cite{kawaguchi}. 
Therefore, 
the error function $L(\boldTheta)$ does not lie in spurious local minima 
in the initial state immediately after the parameter initialization 
under the NTK regime according to Corollary \ref{col-kawa}. 
Therefore, we obtain the following theorem. 
%
\begin{niteth}
\label{thm4} 
Let $q=\rho^{-1}=2^H$. 
Assume that $\boldX\boldX^T$ and $\boldX\boldY^T$ are full rank. 
Then, 
the error function $L(\boldTheta$) does not lie in spurious local minima 
in the initial state of the Kawaguchi model in the NTK regime 
immediately after the parameter initialization. 
\thend
\end{niteth}

\noin
{\bf Analysis during training.}  
Next, we make clear spurious local minima {\it during training} of 
the Kawaguchi model in the NTK regime. 
Jacot et al. theoretically proved that the behavior of neural networks 
during training is described by a related kernel called the neural tangent 
kernel (NTK) in the limit as the widths of the hidden layers tend to infinity \cite{jacot}: 
let $f_{\theta(t)}$ be the network function of a neural network 
which maps an input vector to an output vector where $\theta(t)$ is the vector 
of the parameters of the neural network after the $t$-th learning; 
then, during gradient descent, the dynamics of the network function $f_{\theta(t)}$ 
follows that of the so-called kernel gradient descent in function space 
with respect to a limiting kernel, which only depends on the depth of 
the network, the choice of nonlinearity and the initialization variance. 
More specifically, they proved that for a least-squares regression cost, 
if $f_{\theta(0)}$ is initialized with a normal distribution, 
then the infinite-width limit network function $f_{\theta(t)}$ 
has the normal distribution for all times $t$, and in particular at convergence 
$t \rightarrow \infty$. 
They also made numerical experiments on a ReLU deep neural network 
with a least-squares cost and confirmed that 
the distributions of the network functions are very similar for 
both widths of 50 and 1000: their mean and variance after the 1000th learning 
appear to be close to those of the limiting distribution $t \rightarrow \infty$.

Several researchers also proved that the parameters of deep nonlinear 
neural networks move little from their initial values during training 
such as the stochastic gradient descent in the NTK regime \cite{zhu,du,lee}. 
Especially, Allen-Zhu et al. proved it for deep ReLU neural networks \cite{zhu}. 
Therefore, we realize that the parameters always 
has the same normal distribution as the initial state during training 
of the Kawaguchi model in the NTK regime. 

As a consequence, from Theorem \ref{thm4}, we obtain the following theorem. 
\begin{niteth}
\label{thm5}
Let $q=\rho^{-1}=2^H$. 
Assume that $\boldX\boldX^T$ and $\boldX\boldY^T$ are full rank. 
Then, 
the error function $L(\boldTheta$) does not lie in spurious local minima 
during training of the Kawaguchi model in the NTK regime. 
\thend
\end{niteth}

\section{Discussion}
\label{discuss}
He et al. proposed a weight initialization method for neural networks 
with the ReLU activation function which is commonly used in practice: 
the initial value of each weight between the layer $i$ and the layer $i+1$ 
is set according to either of the normal distributions 
$N ( 0, 2 / n_i )$ or $N ( 0, 2 / n_{i+1} )$ 
independently where $n_i$ is the number of neurons in the layer $i$ \cite{he}. 
This was derived by keeping the variance of 
the net input vector in each layer equal and keeping the variance of the 
back-propagated gradients equal, respectively, 
for the purpose of avoiding their saturation. 
Thus, the learning dynamics of the deep ReLU neural networks 
where the widths of the hidden layers are sufficiently large and 
the parameters are initialized by the He initialization method 
belongs to the NTK regime. 
Therefore, the error function does not lie in spurious local minima in the loss landscape 
of the Kawaguchi model initialized by the He method during training 
in the NTK regime.

Lee et al. theoretically showed that the learning dynamics with 
a certain learning rate 
in parameter space of deep nonlinear neural networks are exactly described 
by a linearized model where the parameters of the deep neural networks 
move little from their initial values in the NTK regime \cite{lee}. 
They did not address the property on local minima in the  NTK regime. 
In contrast, we addressed it in this paper and obtained the result 
that the deep ReLU neural networks do not 
lie in spurious local minima during training under the NTK regime. 
Thus, the result in \cite{lee} and the one of this paper are complementary.

\section{Conclusions}
\label{concl}
We theoretically proved that 
the deep ReLU neural networks 
do not lie in spurious local minima in the loss landscape 
under the Neural Tangent Kernel (NTK) regime, 
that is, in the gradient descent training dynamics of the deep ReLU neural networks 
whose parameters are initialized by a normal distribution 
in the limit as the widths of the hidden layers tend to infinity. 
Especially, the error function does not lie in spurious local minima in the loss landscape 
of the Kawaguchi model initialized by the He initialization method which is commonly used in practice 
during training in the NTK regime. 
The results obtained in this paper approximately hold true for 
a family of ReLU activation functions such as the softplus activation 
function $f(x) = \log(1+\exp(x))$ \cite{dugas}. 
In future studies, we will make clear the property of the local minima 
of deep nonlinear neural networks with the activation functions 
except the ReLU function in the NTK regime.

\subsubsection*{Acknowledgments}
The author would like to give special thanks to 
Dr. R. Karakida, AIST, 
for his valuable comments. 
This work was supported by JSPS KAKENHI Grant Number JP16K00347.




\section*{Supplementary material: \\
Spurious Local Minima of Deep ReLU Neural Networks 
in the Neural Tangent Kernel Regime}

\begin{figure}
\centerline{\epsfig{file=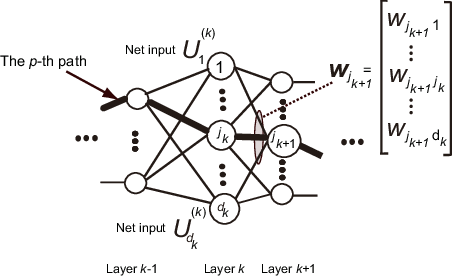,width=9cm,height=5.0cm}}
\caption{Relationship between the weight vector $\boldw_{j_{k+1}}$ and 
the net inputs $U_1^{(k)}, \cdots, U_{d_{k}}^{(k)}$ in Eq. (\ref{eqn4-6}).  }
\label{fig7}
\end{figure}

%
\noin 
{\bf Proof of Theorem \ref{thm2}}. \ 
Take training pattern $i$, output neuron $j$, and path $p$ 
from an input neuron to output neuron $j$ of the Kawaguchi model 
in the NTK regime arbitrarily and fix them. 
Denote by $j_0, \cdots, j_H$ the neurons on path $p$ 
where $j_k$ is the neuron in the $k$-th layer $(k=0, \cdots, H$). 
Let also $w_{j_1 j_0}, \cdots, w_{j_{H+1} j_H }$ denote $H+1$ 
weights on path $p$ 
where $w_{j_{k+1} j_k }$ is the weight between the layer $k$ 
and the layer $k+1$ on path $p$ ($k=0, \cdots, H$) (Fig.~\ref{fig7}). 
Then, for any $\lambda_1, \cdots, \lambda_{H+1} \in \RR$, 
\begin{eqnarray}
\label{eqn4-6}
& & \hspace*{-0.3cm} P \left( [Z_i]_{(j,p)} = 1 \ \Big\vert \ w_{j_1 j_0} = \lambda_1, 
\cdots, w_{j_{H+1} j_H } = \lambda_{H+1} \right) \nonumber\\
&=& \hspace*{-0.3cm} P \left( \boldX_i^T \boldw_{j_1} > 0, \left[ \varphi( U_1^{(1)} ) \cdots \varphi(U_{d_1}^{(1)} ) \right] \boldw_{j_2} > 0, 
%
\cdots, \left[ \varphi( U_1^{(H-1)} ) \cdots \varphi(U_{d_{H-1}}^{(H-1)}) \right] 
\boldw_{j_H} > 0 \right. \nonumber \\
& & \hspace*{5.5cm} \left. \Big\vert \ w_{ j_1 j_0 } = \lambda_1, \cdots, 
w_{ j_{H} j_{H-1} } = \lambda_{H} \right) \ \ 
(\mbox{from Eq.} (\ref{eqn4-2})) \nonumber\\
& & (\mbox{$w_{j_{H+1} j_H} = \lambda_{H+1}$ is removed because it is independent of 
$[Z_i]_{(j,p)}$}) \nonumber\\
&\fallingdotseq& \hspace*{-0.3cm} \frac{1}{2^H}. 
\end{eqnarray}

Here, we can regard that $\lambda_k$ satisfies the inequality 
$\vert \lambda_k \vert \leq 3 \sigma_{k-1} = 1/d_{k-1}$ $(k=1, \cdots, H+1)$ 
because the weight $w_{j_k j_{k-1}}$ obeys the normal distribution 
$N( 0, \sigma_{k-1}^2 )$ where $\sigma_{k-1} = 1/(3d_{k-1})$ 
which means $P(\vert w_{j_k j_{k-1}} \vert \leq 3 \sigma_{k-1} = 1/d_{k-1}) = 0.9973$ 
(the so-called three-sigma rule of thumb). 


We prove below by mathematical induction that the approximate equality $(\fallingdotseq)$ 
in Eq. (\ref{eqn4-6}) holds true.

\noin
[For $H=1$] 
This case corresponds to a three-layered neural network. 
For the sake of simplicity, we let $\boldX_i = (x_1 \cdots x_{j_0} \cdots x_{d_0})^T$ 
and $\boldw_{j_1} = (w_1 \cdots w_{j_0} \cdots w_{d_0})^T$ 
where $w_{j_0} = w_{j_1 j_0}$. Then, 
\begin{eqnarray}
\label{eqn4-7}
& & P \left( \boldX_i^T \boldw_{j_1} > 0 \ \Big\vert \ w_{j_1 j_0} = \lambda_1\right) \nonumber\\
&=& P \left(x_1 w_1 + \cdots + x_{j_0} w_{j_1 j_0} + \cdots + x_{d_0} w_{d_0} > 0 \ 
\Big \vert \ w_{j_1 j_0} = \lambda_1\right) \nonumber\\
&=& P \left(x_1 w_1 + \cdots + x_{j_0} \lambda_1 + \cdots + x_{d_0} w_{d_0} > 0 \ 
\Big \vert \ w_{j_1 j_0} = \lambda_1\right) \nonumber\\
&=& P \Big(x_1 w_1 + \cdots + x_{j_0} \lambda_1 + \cdots + x_{d_0} w_{d_0} > 0 \ \Big) 
\nonumber\\
& & (\mbox{$w_{j_1 j_0} = \lambda_{1}$ is removed because it is independent of 
the other weights}) \nonumber\\
&=& P \Big(x_1 w_1 + \cdots + x_{j_0 - 1} w_{j_0 -1} + x_{j_0 + 1} w_{j_0 + 1} 
+ \cdots + x_{d_0} w_{d_0} > -  x_{j_0} \lambda_1 \ \Big) \nonumber\\
&\fallingdotseq& 
\label{eqn4-7-1}
P \Big(x_1 w_1 + \cdots + x_{j_0 - 1} w_{j_0 -1} + x_{j_0 + 1} w_{j_0 + 1} 
+ \cdots + x_{d_0} w_{d_0} > 0 \ \Big) \\
& & (\mbox{
because $\vert x_{j_0} \lambda_1 \vert \leq 3 \alpha \sigma_0 = \alpha/d_0$ and 
the number of input neurons $d_0$ is sufficiently large}) \nonumber\\
\label{eqn4-7-2}
&=& \frac{1}{2}.  \ \ (\mbox{for the same reason that the last equality 
of Eq. (\ref{eqn4-3}) holds true.}) 
\end{eqnarray}


\noin
[For $H$] 
This case corresponding to a deep ReLU neural network with $H$ hidden layers, 
we show that if the case of $H-1$ holds, then case $H$ also holds. 
Assuming that the case of $H-1$ holds, then
\begin{eqnarray}
\label{eqn4-8}
& & P \left( \boldX_i^T \boldw_{j_1} > 0, \left[ \varphi( U_1^{(1)} ) \cdots \varphi(U_{d_1}^{(1)} ) \right] \boldw_{j_2} > 0, 
%
\cdots, \left[ \varphi( U_1^{(H-1)} ) \cdots \varphi(U_{d_{H-1}}^{(H-1)}) \right] 
\boldw_{j_H} > 0 \right. \nonumber\\
& & \hspace*{8.0cm} \left. \Big\vert \ w_{j_1 j_0} = \lambda_1, 
\cdots, w_{j_H j_{H-1} } = \lambda_H \right) \nonumber\\
%
%
&=& P \left( \left[ \varphi( U_1^{(1)} ) \cdots \varphi(U_{d_1}^{(1)} ) \right] 
\boldw_{j_2} > 0, \cdots, \left[ \varphi( U_1^{(H-1)} ) \cdots \varphi(U_{d_{H-1}}^{(H-1)}) \right] 
\boldw_{j_H} > 0 \ \ \right. \nonumber\\
& & \hspace*{2cm} \left. 
\Big\vert \ \boldX_i^T \boldw_{j_1} > 0, w_{j_1 j_0} = \lambda_1, 
\cdots, w_{j_H j_{H-1} } = \lambda_H \right) \cdot \nonumber\\
& & \hspace*{2cm} P\left( \boldX_i^T \boldw_{j_1} > 0 \ \Big\vert \ w_{j_1 j_0} = \lambda_1, 
\cdots, w_{j_H j_{H-1} } = \lambda_H \right) \nonumber\\
%
%
%
&=& P \left( \left[ \varphi( U_1^{(1)} ) \cdots \varphi(U_{d_1}^{(1)} ) \right] 
\boldw_{j_2} > 0, \cdots, 
%
\left[ \varphi( U_1^{(H-1)} ) \cdots \varphi(U_{d_{H-1}}^{(H-1)}) \right] 
\boldw_{j_H} > 0 \right. \nonumber\\
& & \hspace*{0.5cm} \left. \Big\vert \ \boldX_i^T \boldw_{j_1} > 0, 
\ w_{j_1 j_0} = \lambda_1, \cdots, w_{j_H j_{H-1} } = \lambda_H \right)  
\cdot P\left( \boldX_i^T \boldw_{j_1} > 0 \ \Big\vert \ w_{j_1 j_0} = \lambda_1 \right). \\
& & (\mbox{$w_{j_2 j_1} = \lambda_2, \cdots, w_{j_H j_{H-1}} = \lambda_H$ are removed 
because they are independent of $\boldw_{j_1}$ } ) \nonumber
\end{eqnarray}
Here, the first factor of the right-hand-side of Eq. (\ref{eqn4-8}) 
represents the probability that the path passing through the $H-1$ 
hidden neurons $j_2, \cdots, j_H$ for the input training pattern 
$[ \varphi( U_1^{(1)} ) \cdots \varphi(U_{d_1}^{(1)} ) ]^T \in \RR^{d_1}$ 
such that $U_{j_1}^{(1)} = \boldX_i^T \boldw_{j_1} > 0$ and 
$w_{j_1 j_0} = \lambda_1$ is active. 
Also, for any $1 \leq s \leq d_1$, by denoting $\boldX_i = (x_1 \cdots x_{d_0})^T$ 
and $\boldw_s = (w_1 \cdots w_{d_0})$ for the sake of simplicity, 
$P(\vert \varphi(U_s^{(1)}) \vert \leq \vert U_s^{(1)} \vert 
=\vert \boldX_i^T \boldw_s \vert 
\leq \alpha \sum_{t=1}^{d_0} \vert w_t \vert 
\leq 3\alpha d_0 \sigma_0 = \alpha) = 0.9973 \fallingdotseq 1.0$ 
because each weight $w_t$ obeys the normal distribution $N(0, \sigma^2/{d_0})$ 
which means $P(\vert w_t \vert \leq 3 \sigma_0 = 1/d_0) = 0.9973$ 
(the so-called three-sigma rule of thumb). 
Hence, according to the assumption of mathematical induction, 
the first factor of the right-hand-side of Eq. (\ref{eqn4-8}) is nearly equal 
to $(1/2)^{H-1}$. 
In addition, the second factor of the right-hand-side of Eq. (\ref{eqn4-8}) is 
nearly equal to 1/2 from Eq.(\ref{eqn4-7-2}). 
So, 
\begin{eqnarray}
\label{eqn4-9}
&P& \hspace*{-0.3cm} \left( \boldX_i^T \boldw_{j_1} > 0, \left[ \varphi( U_1^{(1)} ) 
\cdots \varphi(U_{d_1}^{(1)} ) \right] \boldw_{j_2} > 0, 
%
\cdots, \left[ \varphi( U_1^{(H-1)} ) \cdots \varphi(U_{d_{H-1}}^{(H-1)}) \right] 
\boldw_{j_H} > 0 \right. \nonumber\\
& & \hspace*{7.0cm} \left. \Big\vert \ w_{j_1 j_0} = \lambda_1, 
\cdots, w_{j_H j_{H-1} } = \lambda_H \right) \nonumber\\
%
%
&\fallingdotseq& \left(\frac{1}{2}\right)^{H-1} \cdot \frac{1}{2} \nonumber\\
&=& \frac{1}{2^H}, 
\end{eqnarray}
which means that the case of $H$ indeed holds. 
Thus, by mathematical induction, Eq. (\ref{eqn4-6}) holds for any $H \geq 1$. 
Therefore, 
\begin{eqnarray}
\label{eqn4-10}
& & P \left( [Z_i]_{(j,p)} = 1 \ \Big\vert \ w_{j_1 j_0} = \lambda_1, 
\cdots, w_{j_{H+1} j_H } = \lambda_{H+1} \right) \nonumber\\
&\fallingdotseq& \frac{1}{2^H} \hspace*{1cm} \mbox{(from Eq. (\ref{eqn4-6}))}\nonumber\\
&=& P \left( [Z_i]_{(j,p)} = 1 \right). \hspace*{1cm} \mbox{(from Theorem \ref{thm1})}
\end{eqnarray}
This completes the proof. 
\qed 

\noin 
{\bf Proof of Theorem \ref{thm3} }. \ 
Take training pattern $i$, output neuron $j$, path $p$ 
from an input neuron to output neuron $j$ of the deep ReLU neural network 
in the Kawaguchi model in the NTK regime initialized by 
the normal distribution $N( 0, \sigma_l^2 )$ 
where 
$\sigma_l = 1/(3d_l)$ and $d_l$ is the number of neurons in the layer $l$, 
and $\boldmu \in \RR^{d_x}$ arbitrarily and fix them. 
Then, in the same mode of the proof of Theorem \ref{thm1}, it is apparent that 
\begin{equation}
\label{eqn4-12}
P\left([Z_i]_{(j,p)} = 1 \ \big\vert \ \boldX_i = \boldmu \right) = \frac{1}{2^H}. 
\end{equation}
%
Therefore, it follows that 
\begin{eqnarray}
\label{eqn4-13}
P\left([Z_i]_{(j,p)} = 1 \right) &=& \frac{1}{2^H}  \hspace*{1.0cm} (\mbox{from Theorem} \ref{thm1} ) \nonumber\\
&=& P\left([Z_i]_{(j,p)} = 1 \ \ \big\vert \ \ \boldX_i = \boldmu \right) 
\hspace*{0.5cm} (\mbox{from Eq.} (\ref{eqn4-12}) ) 
\end{eqnarray}
holds true. 
Eq. (\ref{eqn4-13}) completes the proof. 
\qed


\begin{thebibliography}{99}
\small{
%
\bibitem{hinton}
Hinton,~G.~E., Osindero,~S., and Teh,~Y. 
``A fast learning algorithm for deep belief nets,'' 
{\it Neural Computation}, 18: 1527-1554, 2006. 

\bibitem{mohamed}
Mohamed,~A-R, Dahl,~G.~E., and Hinton,~G.~E. 
``Deep belief network for phone recognition,'' 
In {\it NIPS Workshop on Deep Learning for Speech Recognition and Related  Applications}, 2009. 

\bibitem{seide}
Seide,~F., Li,~G., and Yu,~D. 
``Conversational speech transcription using context-dependent deep neural networks,'' 
In {\it Proc. Interspeech}, 437-440, 2011. 

\bibitem{parcollet2019}
Parcollet,~T., Ravanelli,~M., Morchid,~M., Linares,~G., 
Trabelsi,~C., Mori,~R.~D., and Bengio.~Y. 
``Quaternion recurrent neural networks,'' 
In {\it International Conference on Learning Representations}, 2019. 





\bibitem{taigman}
Taigman,~Y., Yang,~M., Ranzoto,~M., and Wolf,~L. 
``Deepface: Closing the gap to human-level performance in face verification,'' 
In {\it Proc. Conference on Computer Vision and Pattern Recognition}, 
1701-1708, 2014. 

\bibitem{sutskever}
Sutskever,~I., Vinyals,~O., and Le,~Q.~V. 
``Sequence to sequence learning with neural networks,'' 
In {\it Advances in Neural Information Processing Systems}, 
3104-3112, 2014. 

\bibitem{davies}
Davies,~A., Veliokovic,~P., Buesing,~L. et al. 
``Advancing mathematics by guiding human intuition with AI,'' 
{\it Nature}, vol.600, pp.70-74, 2021. 
https://doi.org/10.1038/s41586-021-04086-x


\bibitem{amari}
Amari,~S., Park,~H., and Ozeki,~T. 
``Singularities affect dynamics of learning in neuromanifolds,'' 
{\it Neural Computation}, vol.18, no.5, pp.1007-1065, 2006. 

\bibitem{cous}
Cousseau,~F., Ozeki,~T., and Amari,~S. 
``Dynamics of learning in multilayer perceptrons near singularities,'' 
{\it IEEE Trans. Neural Networks}, vol.19, no.8, pp.1313-1328, 2008. 

\bibitem{fuku2000} 
Fukumizu,~K. and Amari,~S. 
``Local minima and plateaus in hierarchical structures of multilayer perceptrons,'' 
{\it Neural Networks}, vol.13, no.3, pp.317-327, 2000. 

\bibitem{dauphin}
Dauphin,~Y.~N., Pascanu,~R., Gulcehre,~C., Cho,~K., Ganguli,~S., and Bengio,~Y. 
``Identifying and attacking the saddle point problem in high-dimensional non-convex optimization,'' 
In {\it Advances in Neural Information Processing Systems}, 2933-2941, 2014. 

\bibitem{yun}
Yun,~C., Sra,~S., and Jadbabaie,~A. 
``Small nonlinearities in activation functions create bad local minima in neural networks,'' 
In {\it International Conference on Learning Representations}, 2019. 

\bibitem{choroman-henaff}
Choromanska,~A., Henaff,~M., Mathieu,~M., Arous,~G.~B., and LeCun,~Y. 
``The loss surfaces of multilayer networks,'' 
In {\it Proc. the Eighteenth International
Conference on Artificial Intelligence and Statistics}, 192-204, 2015. 

\bibitem{choroman-lecun}
Choromanska,~A., LeCun,~Y., and Arous,~G.~B. 
``Open problem: the landscape of the loss surfaces of multilayer networks,'' 
In {\it Proc. the 28th Conference on Learning Theory}, 1756-1760, 2015.

\bibitem{kawaguchi}
Kawaguchi,~K. 
``Deep learning without poor local minima,'' 
In {\it Advances in Neural Information Processing Systems 29}, 2016. 

\bibitem{nit2017}
Nitta,~T. 
``Resolution of singularities introduced by hierarchical structure in deep neural networks,'' 
{\it IEEE Trans. Neural Networks and Learning Systems}, vol.28, no.10, pp.2282-2293, 2017. 

\bibitem{laurent}
Laurent,~T. and von Brecht,~J.~H. 
``The multilinear structure of ReLU networks,'' 
In {\it International Conference on Machine Learning}, 2018. 

\bibitem{Liu}
Liu,~B., Liu,~Z., Zhang,~T., Yuan,~T. 
``Non-differentiable saddle points and sub-optimal local minima exist for deep 
ReLU networks,'' 
{\it Neural Networks}, vol.144, pp.75-89, 2021. 

\bibitem{nit2015}
Nitta,~T. 
``Learning dynamics of a single polar variable complex-valued neuron,'' 
{\it Neural Computation}, vol.27, no.5, pp.1120-1141, 2015. 

\bibitem{nit2013}
Nitta,~T. 
``Local minima in hierarchical structures of complex-valued neural networks,'' 
{\it Neural Networks}, vol.43, pp.1-7, 2013. 


\bibitem{jacot}
Jacot,~A., Gabriel,~F., and Hongler,~C. 
``Neural tangent kernel: convergence and generalization in neural networks,'' 
In {\it Advances in Neural Information Processing Systems}, 2018. 

\bibitem{zhu}
Allen-Zhu,~Z., Li,~Y., and Song,~Z. 
``A convergence theory for deep learning via over-parameterization,'' 
In {\it International Conference on Machine Learning}, 2019. 

\bibitem{du}
Du,~S.~S., Lee,~J.~D., Li,~H., Wang,~L., and Zhai,~X. 
``Gradient descent finds global minima of deep neural networks,'' 
In {\it International Conference on Machine Learning}, 2019. 

\bibitem{lee}
Lee,~J., Xiao,~L., Schoenholz,~S.~S., Bahri,~Y., Novak,~R., Sohl-Dickstein,~J., 
and Pennington,~J. 
``Wide neural networks of any depth evolve as linear models under gradient descent,'' 
In {\it Advances in Neural Information Processing Systems}, 2019. 

\bibitem{karakida}
Karakida,~R. and Akaho,~S. 
``Learning curves for continual learning in neural networks: Self-knowledge 
transfer and forgetting,'' 
In {\it International Conference on Learning Representations}, 2022. 


\bibitem{he}
He,~K., Zhang,~X., Ren,~S., and Sun,~J. 
``Delving deep into rectifiers: surpassing human-level performance on 
ImageNet classification,'' 
In {\it Proc. the IEEE International Conference on Computer Vision}, 
1026-1034, 2015. 


\bibitem{dugas}
Dugas,~C., Bengio,~Y., Belisle,~F., Nadeau,~C., and Garcia,~R. 
``Incorporating second-order functional knowledge for better option pricing,'' 
In {\it Advances in Neural Information Processing Systems}, 2000. 


} 

\end{thebibliography}
\end{document}